\documentclass[letterpaper, 10 pt, conference]{ieeeconf}  

\IEEEoverridecommandlockouts                              

\usepackage{cite}

\usepackage{multirow}

\usepackage{graphicx} 
\usepackage{booktabs} 

\usepackage{subcaption}
\usepackage{amsmath}
\usepackage{amssymb}

\usepackage{color,soul}
\usepackage{url}

\title{\LARGE \bf
An LLM-Powered Agent for Physiological Data Analysis: A Case Study on PPG-based Heart Rate Estimation
}

\author{Mohammad Feli$^{1,*}$, Iman Azimi$^{2}$, Pasi Liljeberg$^{1}$, and Amir M. Rahmani$^{2,3}$
\thanks{$^{1}$Department of Computing, University of Turku, Turku, Finland.
        }%
\thanks{$^{2}$Department of Computer Science, University of California, Irvine, USA.}%
\thanks{$^{3}$School of Nursing, University of California, Irvine, USA.}%
\thanks{$^{*}$Corresponding author: Mohammad Feli, {\tt\small mohammad.feli@utu.fi}}%
}

\begin{document}

\maketitle
\thispagestyle{empty}
\pagestyle{empty}

\begin{abstract}
Large language models (LLMs) are revolutionizing healthcare by improving diagnosis, patient care, and decision support through interactive communication. More recently, they have been applied to analyzing physiological time-series like wearable data for health insight extraction. Existing methods embed raw numerical sequences directly into prompts, which exceeds token limits and increases computational costs. Additionally, some studies integrated features extracted from time-series in textual prompts or applied multimodal approaches. However, these methods often produce generic and unreliable outputs due to LLMs' limited analytical rigor and inefficiency in interpreting continuous waveforms. In this paper, we develop an LLM-powered agent for physiological time-series analysis aimed to bridge the gap in integrating LLMs with well-established analytical tools. Built on the OpenCHA, an open-source LLM-powered framework, our agent powered by OpenAI’s GPT-3.5-turbo model features an orchestrator that integrates user interaction, data sources, and analytical tools to generate accurate health insights. To evaluate its effectiveness, we implement a case study on heart rate (HR) estimation from Photoplethysmogram (PPG) signals using a dataset of PPG and Electrocardiogram (ECG) recordings in a remote health monitoring study. The agent’s performance is benchmarked against OpenAI GPT-4o-mini and GPT-4o, with ECG serving as the gold standard for HR estimation. Results demonstrate that our agent significantly outperforms benchmark models by achieving lower error rates and more reliable HR estimations. The agent implementation is publicly available on GitHub\footnote{https://github.com/mohammadfeli/CHA-PPGHR}.
\end{abstract}

\section{INTRODUCTION}

Large language models (LLMs) have emerged as transformative tools in healthcare, reshaping the interaction between medical professionals, patients, and medical data \cite{yang2023large}. Their ability to understand and generate human-like text has led to widespread adoption across various applications, including medical diagnosis, patient care, clinical decision support, drug discovery, health recommendation systems, and healthcare chatbots \cite{nazi2024large}. By enabling personalized and efficient communication, LLMs provide more interactive and proactive healthcare solutions. They are typically adapted using prompt engineering,  which involves crafting precise input prompts to refine responses, and fine-tuning, which customizes the model for specific healthcare applications \cite{peng2024model}.

Recently, several LLM-based agents have been developed to analyze medical time series and physiological signals \cite{chan2024medtsllm}. These agents have been applied to analyze physiological data—such as those collected from wearable devices—and derive insights from complex sensor data. They aim to use LLMs’ reasoning capabilities to perform analytical tasks by constructing structured prompts from medical time series.

To integrate LLMs with physiological time-series data, three primary strategies have been explored in the literature. The first approach involves directly incorporating numerical time-series sensor data into textual prompts to perform various analytical tasks using LLMs \cite{liu2023large, kim2024health, chan2024medtsllm, imran2024llasa, bohi2024large}. However, these methods struggle with handling large numerical sequences, often exceeding token limits in LLMs and increasing computational costs. The second approach extracts features from time-series health data and integrates them into textual prompts \cite{cosentino2024towards, fang2024physiollm, liu2024large, yu2023zero}. While this reduces token overhead, it still leads to generic and unreliable outputs, due to LLMs' limitations in analytical reasoning. The third approach, multimodal prompting, introduces visual representations of physiological signals, such as Electrocardiogram (ECG) and Photoplethysmogram (PPG), for LLM-based biosignal analysis \cite{yoon2024my, tang2023alpha}. Similar to feature-based prompting, these methods also fail to provide analytically grounded insights, as LLMs are not inherently designed to interpret continuous waveforms.

We believe that there is a critical research gap in existing studies regarding the integration of LLMs with reliable analytical tools for physiological time-series analysis. Rather than using LLMs for direct numerical reasoning, their strengths in strategic planning, critical thinking, and interaction can be leveraged. LLMs can serve as intelligent intermediaries, facilitating seamless user interaction with well-established healthcare analysis tools. Moreover, by linking LLMs to user-specific health data sources, they are able to deliver context-aware, personalized insights. This approach shifts the role of LLMs from analysis to orchestration,  enabling the creation of agents that deliver interactive, interpretable, and reliable analytical solutions.

In this paper, we develop an LLM-powered agent designed to analyze the user's input query, perform physiological time-series data analysis methods, and generate responses accordingly. The agent interacts with users by receiving health analysis prompts, accessing user-specific data, leveraging AI and analytical models, and generating accurate health insights. Leveraging the OpenCHA framework \cite{abbasian2023conversational}, our agent features an orchestrator that coordinates interactions between a user interface, data sources, and external analytical tools for advanced physiological data processing. To demonstrate its effectiveness, we implement a case study on heart rate (HR) estimation from PPG signals. We evaluate the proposed agent using a dataset comprising PPG and ECG data from 46 individuals collected in a remote health monitoring study, where ECG serves as the gold standard for HR estimation. Additionally, we assess the agent performance in comparison to the OpenAI GPT-4o and GPT-4o-mini models \cite{openai2024}. To encourage reproducibility and further research, we have publicly released the implementation of our agent on GitHub\footnote{https://github.com/mohammadfeli/CHA-PPGHR}.

\section{Related Works}
In this section, we review state-of-the-art methods for LLM-based physiological time-series data analysis. We categorize existing approaches into three main groups: (i) direct integration of raw time-series sensor data into textual prompts, (ii) textual prompting using extracted features from time-series data, and (iii) multimodal prompting, which leverages visual representations of physiological signals. 

Studies utilizing textual prompting with direct integration of time-series data represent the numerical data in a structured text format for the model. One of the earliest studies in this context \cite{liu2023large} demonstrated the use of LLMs for tasks such as arrhythmia detection and physical activity recognition by directly integrating interbeat interval sequences and accelerometer sensor data into prompts. Similarly, Kim et al. \cite{kim2024health} explored LLMs for health prediction by incorporating various sensor readings into queries, such as step count, calories burned, and resting heart rate. Chan et al. \cite{chan2024medtsllm} proposed an LLM-based framework for tasks such as semantic segmentation, boundary detection, and anomaly detection, embedding biosignals directly into model inputs. Additionally, Imran et al. \cite{imran2024llasa} employed LLMs for physical activity analysis, where gyroscope and accelerometer sensor data were incorporated into prompts. LLMs have also been applied to cardiac activity monitoring, metabolic health prediction, and sleep detection by leveraging wearable sensor data  \cite{bohi2024large}. However, LLMs are not inherently specialized for processing long raw numerical sequences. Physiological time-series data often consist of thousands or even millions of data points, and representing them as textual input for LLMs can result in poor performance. This approach also exceeds the token limits of most LLMs and leads to high computational costs due to the large number of tokens. For instance, the input token limit of GPT-4 models is 128,000 tokens, while a 16-minute PPG signal sampled at just 20 Hz results in over 130,000 tokens.

To address these limitations, some studies have instead extracted numerical features from medical time-series data and integrated them into textual prompts. Cosentino et al. \cite{cosentino2024towards} fine-tuned the Gemini models \cite{team2023gemini} for sleep and fitness monitoring by incorporating physiological parameters extracted from wearable data into prompts. Similarly, PhysioLLM \cite{fang2024physiollm} was introduced to generate personalized health insights using LLMs that integrate wearable-derived features with contextual information. An LLM-based blood pressure estimation model \cite{liu2024large} was also developed, integrating extracted physiological features from ECG and PPG signals. Furthermore, Yu et al. \cite{yu2023zero} proposed an arrhythmia detection approach that incorporates physiological and morphological features extracted from ECG signals. While these methods address the challenge of excessive token usage in prompting, they still suffer from generic and unreliable outputs. Since LLMs are inherently designed to process and generate human-like text by tokenizing input data and predicting subsequent tokens, their responses—despite advanced prompt engineering or fine-tuning—often lack analytical rigor and reliability, which is required for robust physiological data interpretation.

More recently, multimodal prompting has been explored as an alternative approach by incorporating visual representations of physiological signals \cite{yoon2024my, tang2023alpha}. Yoon et al. \cite{yoon2024my} proposed a visual prompting approach, providing the model with images of signals such as accelerometer, ECG, Electromyogram (EMG), and respiration data for tasks including human activity recognition, arrhythmia diagnosis, hand gesture recognition, and stress detection. Similarly, Tang et al. \cite{tang2023alpha} applied visual prompting by providing the model with PPG signal images for heart rate estimation. However, similar to feature-based prompting methods, these multimodal approaches also suffer from generating generic responses rather than analytically grounded insights, as LLMs are not inherently designed to directly interpret waveforms from continuous signals.

We believe the following challenges in the literature need to be addressed:
\begin{enumerate}
    \item Existing LLM-based health time-series analysis methods struggle with handling large numerical sequences, often exceeding LLM token limits, leading to high computational costs and performance degradation.
    \item LLMs are not inherently designed for numerical reasoning, resulting in generic and analytically unreliable outputs when utilized for direct time-series analysis in healthcare applications.
\end{enumerate}




\section{Method}
We develop an LLM-powered agent to respond to users with reliable insights from physiological data. This agent is designed to interact with users by receiving prompts for extracting health insights from data, incorporating user-specific data sources, leveraging AI and analysis models, and delivering analytically grounded results. To achieve this, we employ an open-source agent-based framework \cite{abbasian2023conversational} as the backbone of our system. Our agent includes an orchestrator that coordinates interactions between multiple components, including a user interface, user data sources, and analytical models. Moreover, it sends the analysis results to an LLM, allowing the model to contextualize the findings with its internal knowledge and generate user-friendly responses. 

In the following, we first describe the architecture of our agent, detailing its different components and workflow. We then present a case study demonstrating its implementation for HR extraction from PPG signals.

\subsection{Architecture}
\begin{figure}[!t]
    \centering
    \includegraphics[width=0.90\columnwidth]{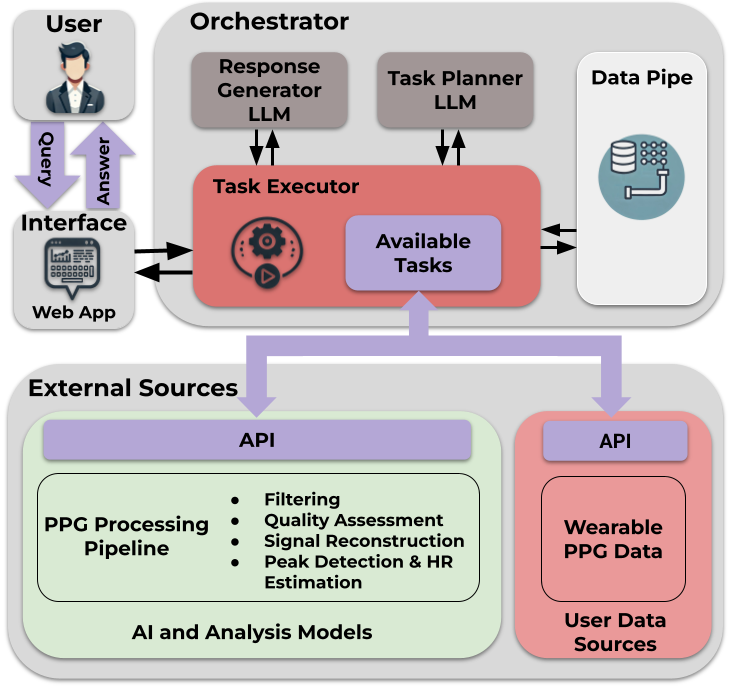} 
    \caption{Proposed LLM-based agent framework for physiological time-series data analysis: case study on HR estimation from PPG.}
    \label{fig:arch-diagram}
\end{figure}

Our LLM-based agent framework consists of three main components: Interface, Orchestrator, and External Sources (see Figure \ref{fig:arch-diagram}), which are outlined below.

\subsubsection{Interface} 
The Interface serves as the primary communication bridge between users and the framework, enabling interactions through text-based queries. Users can submit requests for time-series analysis, which the Interface forwards to the Orchestrator for processing. Additionally, it delivers results in a structured, user-friendly format while allowing users to refine or expand their queries as needed.

\subsubsection{Orchestrator} 
The Orchestrator serves as the core component of our agent, responsible for problem-solving, strategic planning, task execution, and generating analytical responses based on user queries. It acts as the central decision-making unit, coordinating interactions between the Interface, External Sources, and LLM to extract meaningful insights from physiological data. The Orchestrator analyzes user prompts, retrieves relevant time-series data from sources, executes processing tasks, and synthesizes the final response, ensuring the delivery of accurate health insights.

The Orchestrator consists of four key components: 1) \textit{Task Planner}: Decomposes user queries into executable tasks, enabling strategic planning and decision-making. 2) \textit{Task Executor}: Executes planned tasks, including retrieving users' time-series data and performing analysis. 3) \textit{Data Pipe}: Serves as a repository for acquired metadata and intermediate data for efficient processing. 4) \textit{Response Generator}: Synthesizes analytical results, integrates them with LLM reasoning, and delivers a structured response. 

The Orchestrator handles failures (e.g., in data retrieval or task execution) through iterative communication between the Task Planner and Task Executor. When a step fails, the orchestrator can dynamically replan by revisiting earlier stages or prompting the user for clarification. Additionally, the agent is not limited to specific predefined tasks. OpenCHA’s modular design enables the orchestrator to invoke any domain-specific tools based on the user’s query. This design ensures flexibility and extensibility to handle a broader range of user queries by simply registering additional analytical tools as new tasks within the framework.

For strategic task planning in Orchestrator, we employ the Tree of Thought prompting technique \cite{yao2024tree}, which allows the LLM to generate multiple task execution strategies, evaluate their advantages and limitations, and select the most effective approach for a given query. Additionally, we use OpenAI’s GPT-3.5-turbo model \cite{openai2024} as the base LLM in our implementation. 

\subsubsection{External sources} 
External sources play a crucial role in accessing external data and processing tools. Our agent integrates two primary external sources: (1) user data sources and (2) AI and analysis models (see Figure \ref{fig:arch-diagram}).

\subsection{Case study: HR Estimation from PPG}
To demonstrate the effectiveness of our LLM-powered agent, we implement a case study for HR estimation from PPG signals. We design two components as external sources: (1) a PPG dataset collected by wearables and (2) an established processing pipeline for HR extraction from PPG. The details of these two components are outlined below.

\subsubsection{Wearable PPG data source}\label{dataset}
This component provides the agent with users' PPG data for analysis. When a user requests signal analysis, the agent executes a task that retrieves the corresponding signal from this data source. This ensures access to relevant time-series data for a specific user. To achieve this, we utilize a PPG dataset from a remote health monitoring study \cite{mehrabadi2020sleep}. In this study, participants wore a Samsung Gear Sport watch {\cite{SamsungGearSport}} on their non-dominant hand and a Shimmer3 device {\cite{Shimmer3ECGUnit}} on their chest, continuously recording PPG and ECG signals. The data were collected over a 24-hour monitoring period in free-living conditions, capturing participants' daily routines and activities.

The study was conducted in southern Finland between July and August 2019, involving 46 participants (23 men and 23 women). To be eligible, participants needed to be in good health with no cardiovascular diseases, aged 18 to 55, capable of using wearable devices during work, and without physical activity restrictions. The study's purpose was explained to participants in face-to-face meetings before enrollment. Table {\ref{tab:info}} provides an overview of participant demographics, based on 42 individuals, as data from four participants is unavailable.

\begin{table}[!t]
\scriptsize
\centering
\caption{Participants’ background information in PPG dataset.}
\label{tab:info}
\begin{tabular}{lll}
\toprule
\textbf{Characteristic}                   & \textbf{Type}                       & \textbf{Values} \\ \hline
                                          & Men                                 & 33 (6)          \\
\multirow{-2}{*}{Age, mean (SD)}  & Women                               & 31.5 (6.6)      \\ \hline
                                          & Men                                 & 24.4 (5.6)      \\
\multirow{-2}{*}{BMI, mean (SD)}          & Women                               & 25.5 (2.9)      \\ \hline
                                          & Almost daily & 12 (27)         \\
                                          & Once a week                         & 9 (20)          \\
\multirow{-3}{*}{Physical activity, n (\%)} & $>$Once a week                        & 21 (47)         \\ \hline
                                          & Working                             & 32 (71)         \\
                                          & Unemployed                          & 1 (2)           \\
                                          & Student                             & 8 (18)          \\
\multirow{-4}{*}{Employment status, n(\%)}    & Other                               & 1 (2)           \\ \toprule
\end{tabular}
\end{table}

The Samsung Gear Sport watch {\cite{SamsungGearSport}} was used to collect PPG signals from the wrist. Designed for long-term data collection, the watch is lightweight, waterproof, and measures 44.6 × 42.9 × 11.6 mm, weighing 67 g, including the strap. It operates on the open-source Tizen system and has a three-day battery life. The device features optical and inertial measurement unit (IMU) sensors, with PPG signals recorded at a sampling rate of 20 Hz.

The Shimmer3 device {\cite{Shimmer3ECGUnit}} was used to capture ECG data from the chest. This portable system has ample storage and battery capacity, enabling continuous 24-hour data collection. The device recorded 12-channel ECG signals using four electrodes on the torso at a sampling frequency of 512 Hz. In our analysis, we used only Lead II ECG signals as the ground truth for HR calculations.

Participants were instructed to wear the Samsung Gear Sport on their non-dominant hand and the Shimmer3 device on their chest for a full day. PPG signals were collected in 16-minute intervals every 30 minutes, while ECG data were recorded continuously for 24 hours during daily activities. To ensure accurate data calibration, the first and last minutes of each recording were discarded from the analysis.

\textbf{Ethics}: This study was conducted in compliance with the Declaration of Helsinki and the Finnish Medical Research Act (\#488/1999). Ethical approval was obtained from the Ethics Committee for Human Sciences at the University of Turku (Statement \#44/2019). Participants were fully informed about the study both orally and in writing before providing written consent. Participation was voluntary, with the right to withdraw at any time without explanation.

\subsubsection{PPG processing pipeline}
This component provides the data analysis models for HR estimation from PPG signals. When a user requests HR estimation, the agent retrieves the relevant signal from the PPG data source and then executes this task for analysis. PPG signals are highly susceptible to motion artifacts and noise, particularly in free-living conditions \cite{orphanidou2017signal}. To ensure reliable HR estimation, we utilize a robust PPG processing pipeline, previously proposed and validated in prior research \cite{feli2023end}.

The pipeline consists of multiple processing stages:
\begin{itemize}
    \item \textbf{Filtering}: A high-pass filter with a 0.5 Hz cut-off frequency is applied to remove low-frequency noise from the signal. Since raw PPG signals collected in free-living environments are prone to various interferences, this step eliminates unwanted low-frequency components that do not contribute to HR estimation.

    \item \textbf{Signal quality assessment}: This stage incorporates a well-established model \cite{feli2023energy} to distinguish clean PPG signals from noisy or artifact-affected segments. Motion artifacts can significantly distort the waveform, making it difficult to extract reliable HR values. Therefore, assessing signal quality is essential for selecting high-quality segments for analysis.

    \item \textbf{Signal reconstruction}: In cases where the PPG signal contains minor noise interference, the corrupted sections are reconstructed. Given the quasi-periodic nature of PPG signals, the missing or noisy portions can be inferred based on adjacent clean segments. A deep convolutional generative adversarial network (GAN)-based approach \cite{wang2022ppg} is utilized to reconstruct noisy signal segments, supporting corrections of up to 15 seconds of corrupted data.

    \item \textbf{Peak detection and HR estimation}: A deep-learning-based PPG peak detection algorithm \cite{kazemi2022robust} is applied to identify systolic peaks, which are critical for deriving vital signs from PPG waveforms. Once the peaks are identified, HR is computed in beats per minute (BPM).
\end{itemize}

\section{Evaluation and Results}
We evaluate our LLM-powered agent using the dataset described in Section \ref{dataset}. For this evaluation, we use PPG data from 12 participants, comprising 621 PPG recordings, each spanning 15 minutes. These 12 participants were held out as a test set and were not used in any stage of model development or fine-tuning in the PPG processing pipeline. The remaining 34 participants' data in the dataset were used exclusively for training and validation of the machine learning models in the pipeline. The pipeline was fully finalized before evaluating the 12 held-out participants to ensure an unbiased assessment of the agent's performance.

\subsection{Benchmark Models for Comparison} 
We used OpenAI’s GPT-4o-mini and GPT-4o \cite{openai2024} as benchmark models. GPT-4o-mini is a lighter, more cost-efficient version of GPT-4o, designed for faster response times while maintaining strong reasoning capabilities. GPT-4o is one of the latest OpenAI’s model, offering state-of-the-art performance in natural language understanding, reasoning, and multimodal processing. Both models were utilized via the OpenAI API \cite{OpenAIPlatform2024}, allowing us to efficiently process all samples in our evaluation.  

For implementing these OpenAI models, we utilized OpenAI Assistants \cite{OpenAI_Assistants_Overview}, which enhance the analytical rigor of LLM responses by generating and executing code instead of relying solely on text-based reasoning. Assistants provide built-in tools, including the Code Interpreter and File Search, to facilitate data analysis. The Code Interpreter allows Assistants to write and execute Python code in a sandbox environment, enabling the processing of data files and iterative problem-solving. If an initial code execution fails, the Assistant can modify and rerun the code until it succeeds. The File Search tool enables Assistants to access user-provided data files, with OpenAI automatically parsing and preparing the files for analysis. This approach ensures that responses are derived from analytical processes rather than generic text-based outputs.

\subsection{Gold reference and comparison criteria}
We evaluate the HR values extracted from PPG using our LLM-powered agent and OpenAI models, comparing them against ECG-derived HR values as the gold reference. Ground truth HR labels are obtained from ECG signals using the Elgendi et al. method \cite{elgendi2010frequency}.

For evaluation, we compute Mean Absolute Error (${MAE} = \frac{1}{n} \sum_{i=1}^{n} |y_i - \hat{y}_i|$), Root Mean Square Error (${RMSE} = \sqrt{\frac{1}{n} \sum_{i=1}^{n} (y_i - \hat{y}_i)^2}$), Mean Absolute Percentage Error (${MAPE} = \frac{100\%}{n} \sum_{i=1}^{n} \left| \frac{y_i - \hat{y}_i}{y_i} \right|$), and Median Absolute Difference (${MAD} = \text{median}(|y_i - \hat{y}_i|)$), where $n$ is the number of samples, $y_i$ represents the ground-truth values, and $\hat{y}_i$ is the predicted value of HR. Additionally, we perform linear regression analysis to assess the linear relationship between estimated and reference HR values and use the Bland-Altman method \cite{bland1986statistical} to evaluate the agreement between the estimated and ground truth HR values.

\subsection{Results and Performance Analysis} 
\begin{figure}[!t]
  \centering
  \begin{subfigure}[b]{0.49\textwidth}
    \includegraphics[width=0.98\columnwidth]{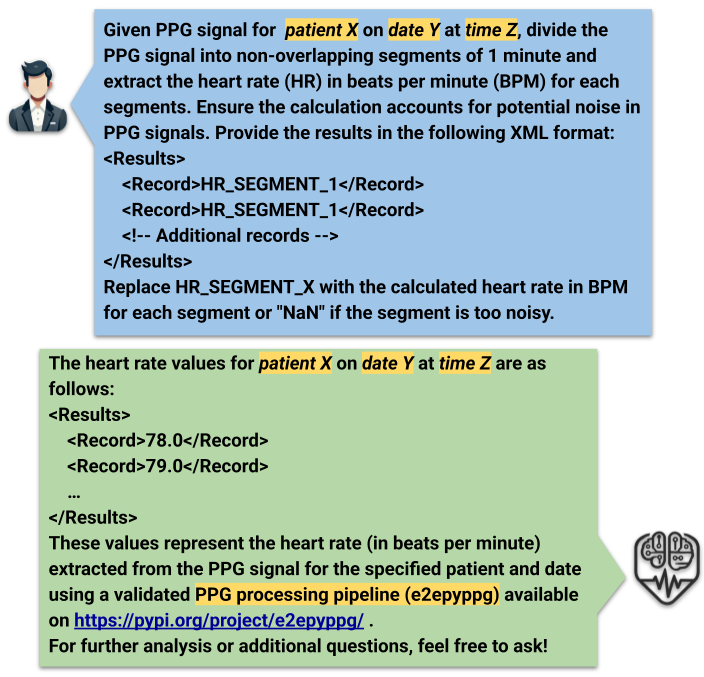}
    \caption{Proposed agent}
    \label{fig:agent-res}
  \end{subfigure}
  \hfill 
  \begin{subfigure}[b]{0.49\textwidth}
    \includegraphics[width=0.98\columnwidth]{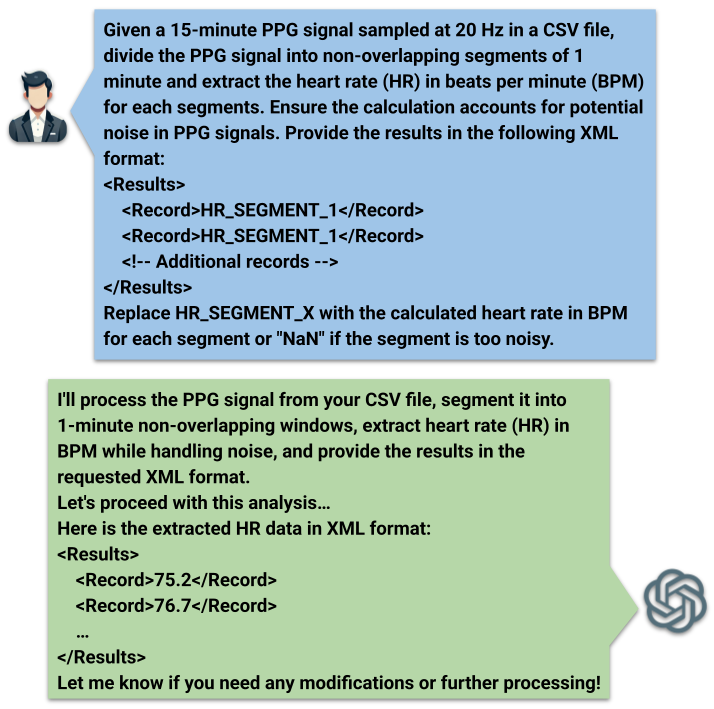}
    \caption{OpenAI model}
    \label{fig:openAI-res}
  \end{subfigure}
  \caption{A sample question and response from the proposed agent and OpenAI model.}
  \label{fig:responses}
\end{figure}

In this section, we present the evaluation results for our LLM-powered agent and the benchmark OpenAI models in HR estimation from PPG. Figure \ref{fig:responses} illustrates sample prompts and responses from our agent and the OpenAI models.

As demonstrated in the prompt, our LLM-powered agent is tasked to extract HR from a user’s PPG signal by specifying the user ID, date, and time. Since the agent is integrated with the user’s PPG data source, it can automatically retrieve the relevant PPG signal and process it for HR estimation. Additionally, the agent explicitly references the analysis method used, which in this case is our PPG processing pipeline.

In contrast, when using the OpenAI models, the PPG signal must be manually uploaded along with the prompt. Unlike our agent, these models do not specify a reference for the analysis method, as the HR estimation is derived from code generated and executed within OpenAI Assistants. Despite using nearly identical prompts, the difference lies in how the data is accessed. Moreover, both our agent and the OpenAI models were instructed to handle potential noise in the PPG signals as part of the prompt. To extract the HR values from the responses, we utilized regular expressions to identify and retrieve HR values enclosed within XML tags. 

Table \ref{tab:errors} presents the pairwise error metrics comparing HR values extracted from PPG using our agent and OpenAI models against the refrence ECG HR values. As shown, our agent achieves the lowest error rates, with an MAE of 2.83, RMSE of 5.47, MAPE of 0.04, and MAD of 1.90. Additionally, GPT-4o exhibited slightly better performance than GPT-4o-mini. These results demonstrate that our agent significantly outperforms the OpanAI models across all error evaluation metrics, showing its superior accuracy.

\begin{table}[!t]
\centering
    \caption{Comparison of the proposed agent and benchmark models in HR Estimation from PPG.}
    \begin{tabular}{lllll}
    \hline
        \textbf{Method}   & \textbf{MAE}  & \textbf{RMSE} & \textbf{MAPE} & \textbf{MAD}  \\ \hline
        GPT-4o-Mini       & 9.41          & 16.47         & 0.13          & 4.62          \\
        GPT-4o           & 8.93          & 14.57         & 0.12          & 4.54          \\
        \textbf{Proposed} & \textbf{2.83} & \textbf{5.47} & \textbf{0.04} & \textbf{1.90} \\ \hline
    \end{tabular}
    \label{tab:errors}
\end{table}

\begin{figure}[!t]
  \centering
  \begin{subfigure}[b]{0.36\textwidth}
    \includegraphics[width=\textwidth, trim=0cm 0.5cm 0cm 0.5cm, clip]{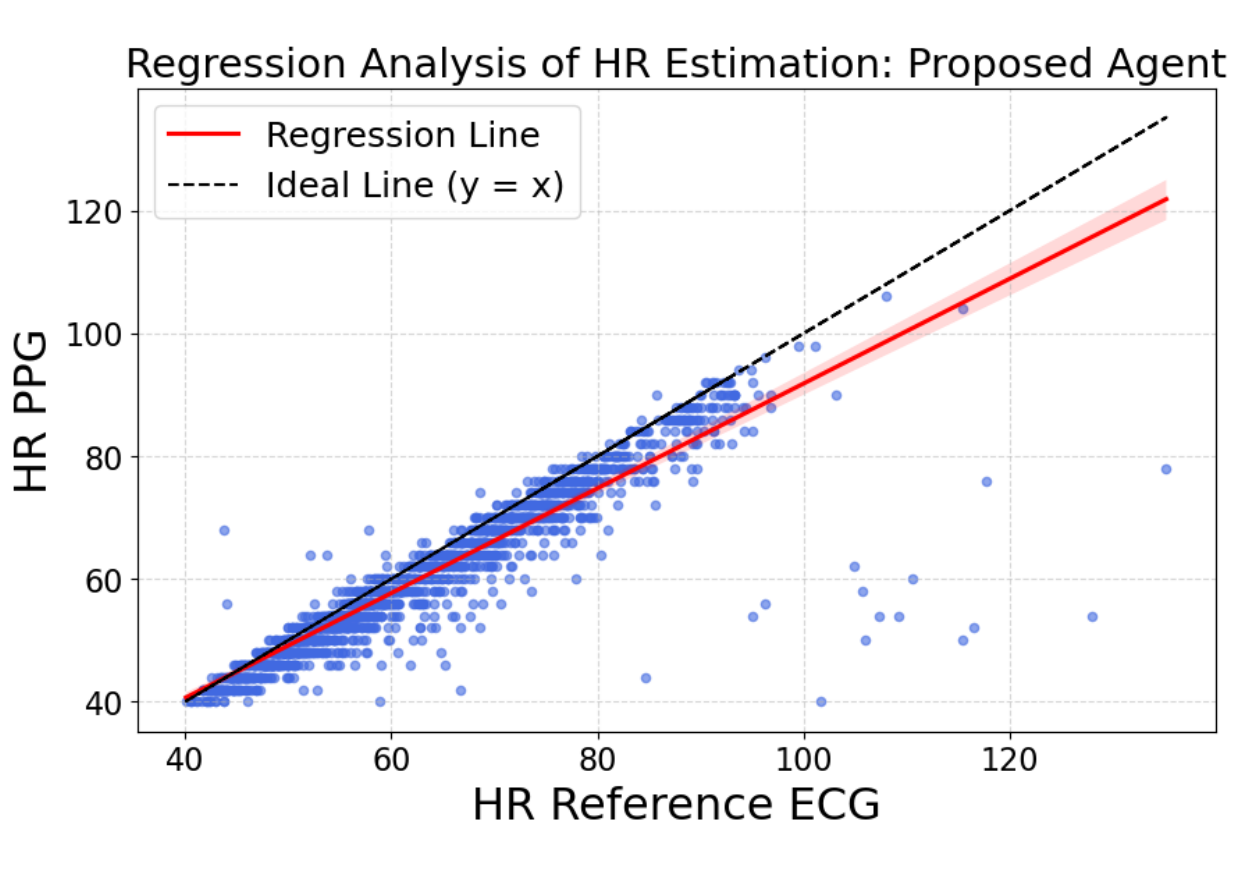}
    \caption{Proposed agent}
    \label{fig:reg-proposed}
  \end{subfigure}
  \hfill 
  \begin{subfigure}[b]{0.36\textwidth}
    \includegraphics[width=\textwidth, trim=0cm 0.5cm 0cm 0.5cm, clip]{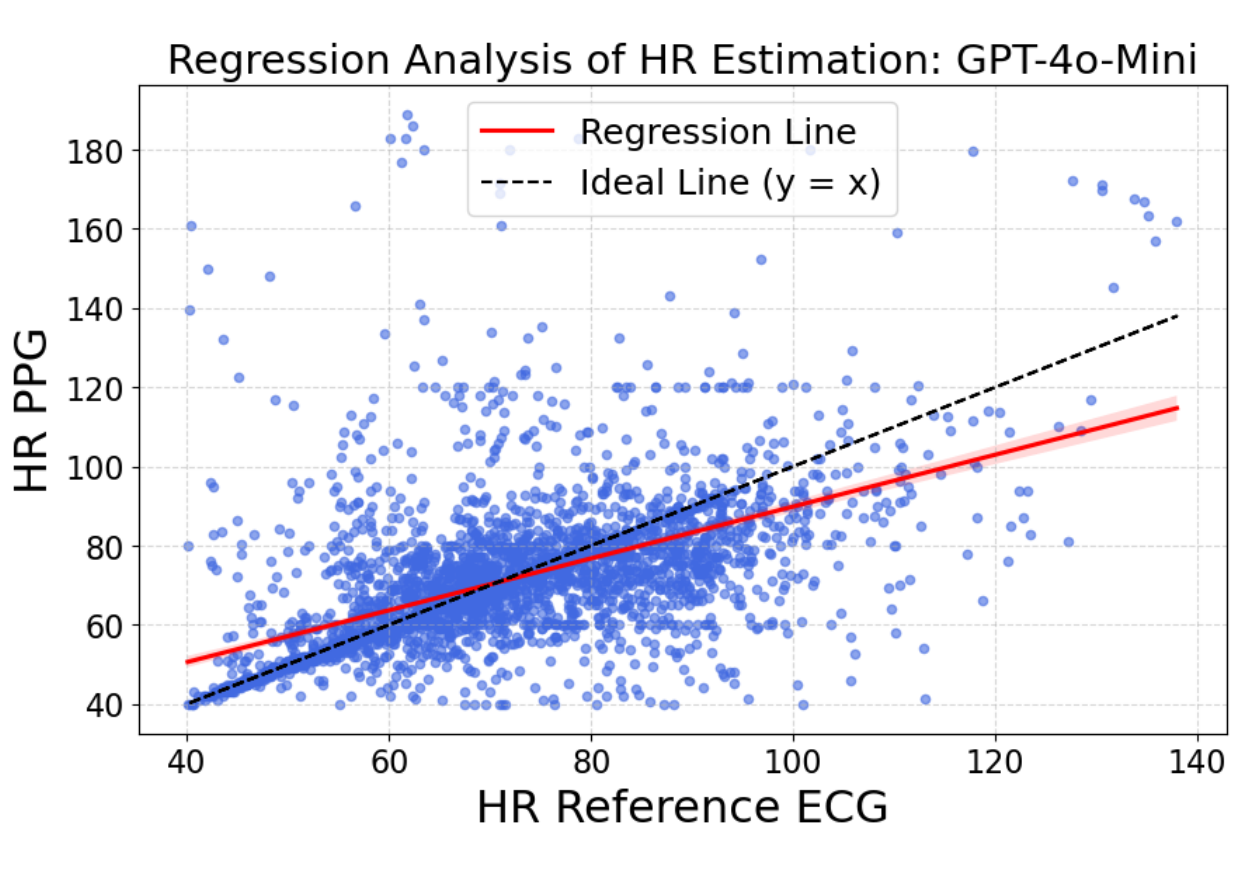}
    \caption{GPT4o-Mini}
    \label{fig:reg-4o-mini}
  \end{subfigure}
  \hfill 
  \begin{subfigure}[b]{0.36\textwidth}
    \includegraphics[width=\textwidth, trim=0cm 0.5cm 0cm 0.5cm, clip]{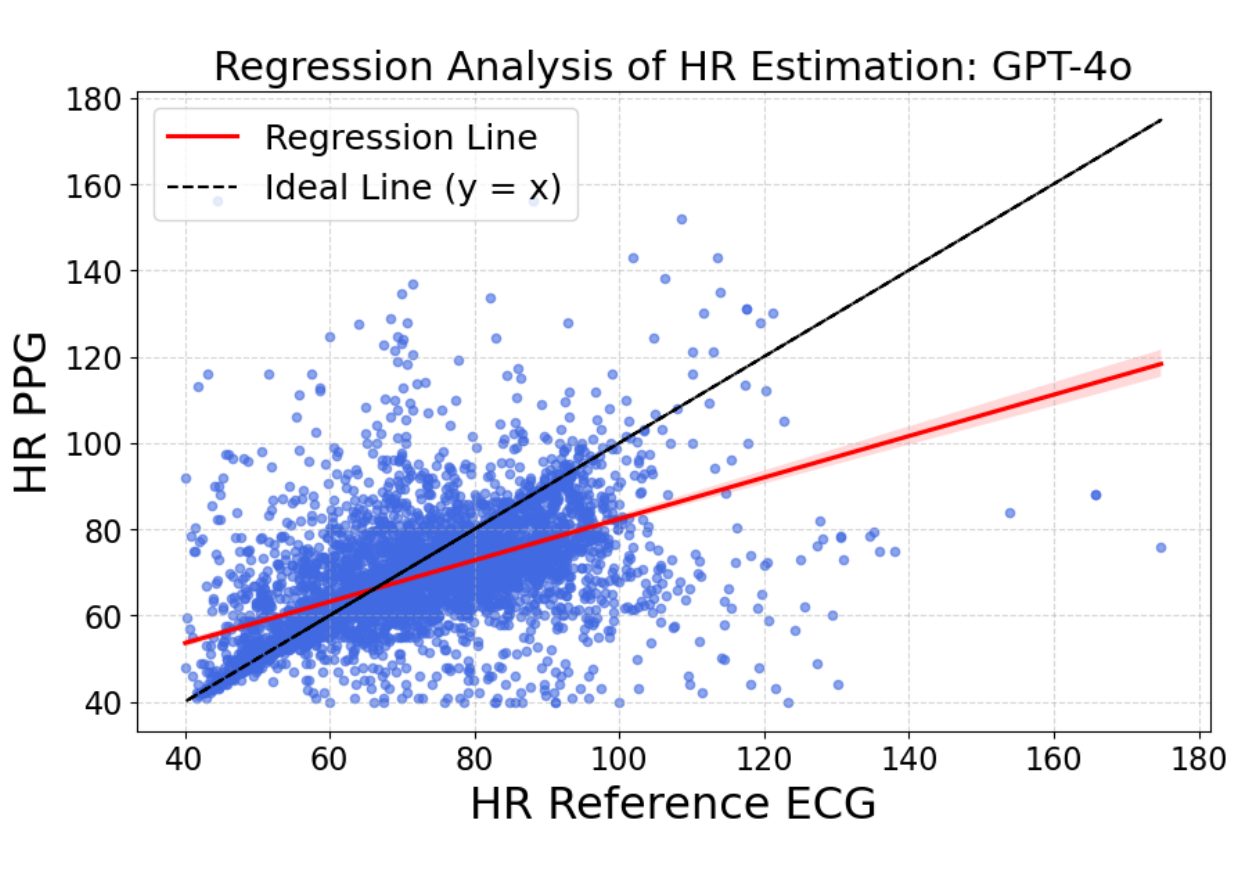}
    \caption{GPT4o}
    \label{fig:reg-4o}
  \end{subfigure}

  \caption{Regression analysis plots of the proposed agent and benchmark OpenAI models}
  \label{fig:reg-plots}
\end{figure}

Figure \ref{fig:reg-plots} illustrates the regression analysis results comparing HR values extracted by our agent and OpenAI models against the reference ECG HR values. In the plots, the red line represents the regression fit, while the black line indicates the optimal y = x relationship, where the estimated and actual values are identical. As shown, the regression line for our agent's estimations closely aligns with the ideal line, indicating a strong correlation with the reference values. In contrast, the regression lines for GPT-4o-Mini and GPT-4o deviate more significantly, showing higher estimation errors in comparison to our agent.

\begin{figure}[!t]
  \centering
  \begin{subfigure}[b]{0.36\textwidth}
    \includegraphics[width=\textwidth, trim=0cm 0.5cm 0cm 0.5cm, clip]{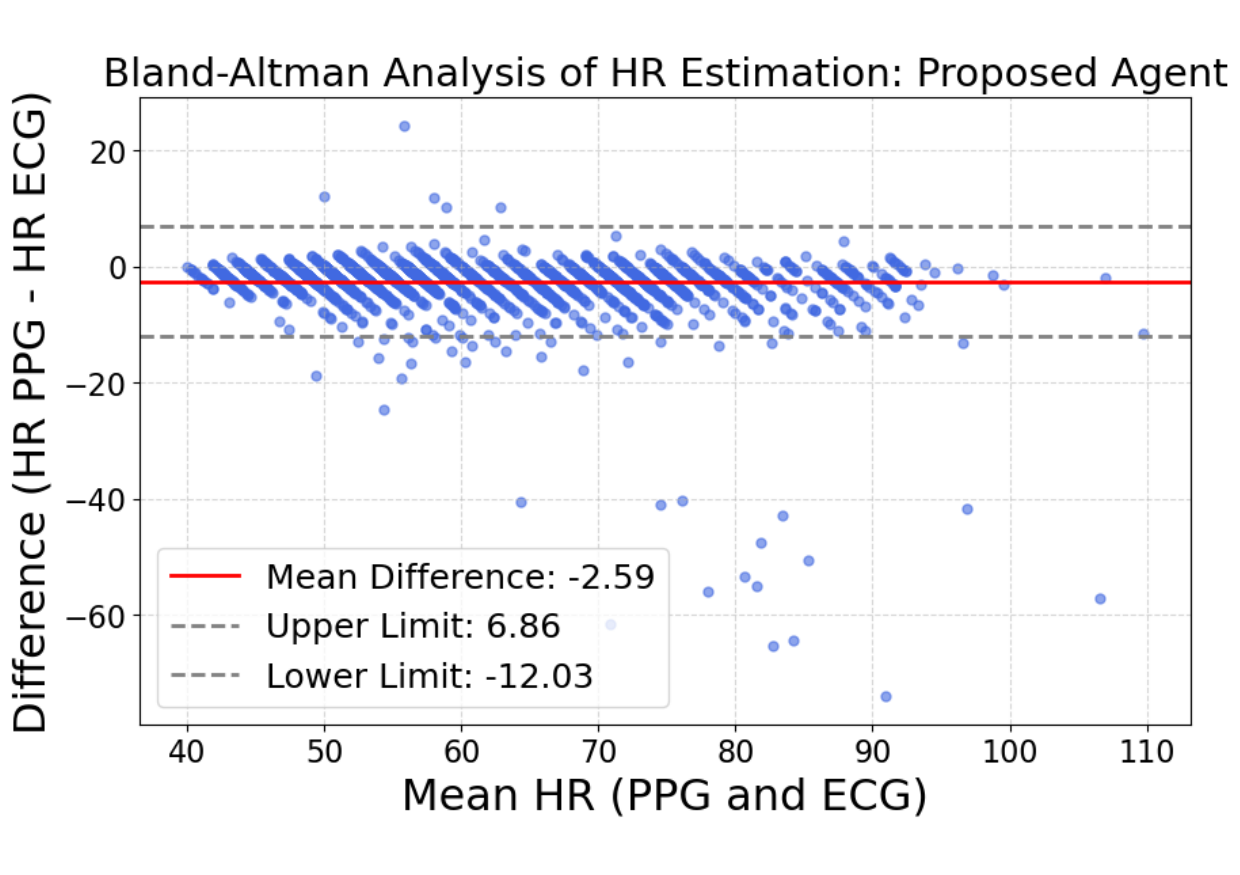}
    \caption{Proposed agent}
    \label{fig:bland-proposed}
  \end{subfigure}
  \hfill 
  \begin{subfigure}[b]{0.36\textwidth}
    \includegraphics[width=\textwidth, trim=0cm 0.5cm 0cm 0.5cm, clip]{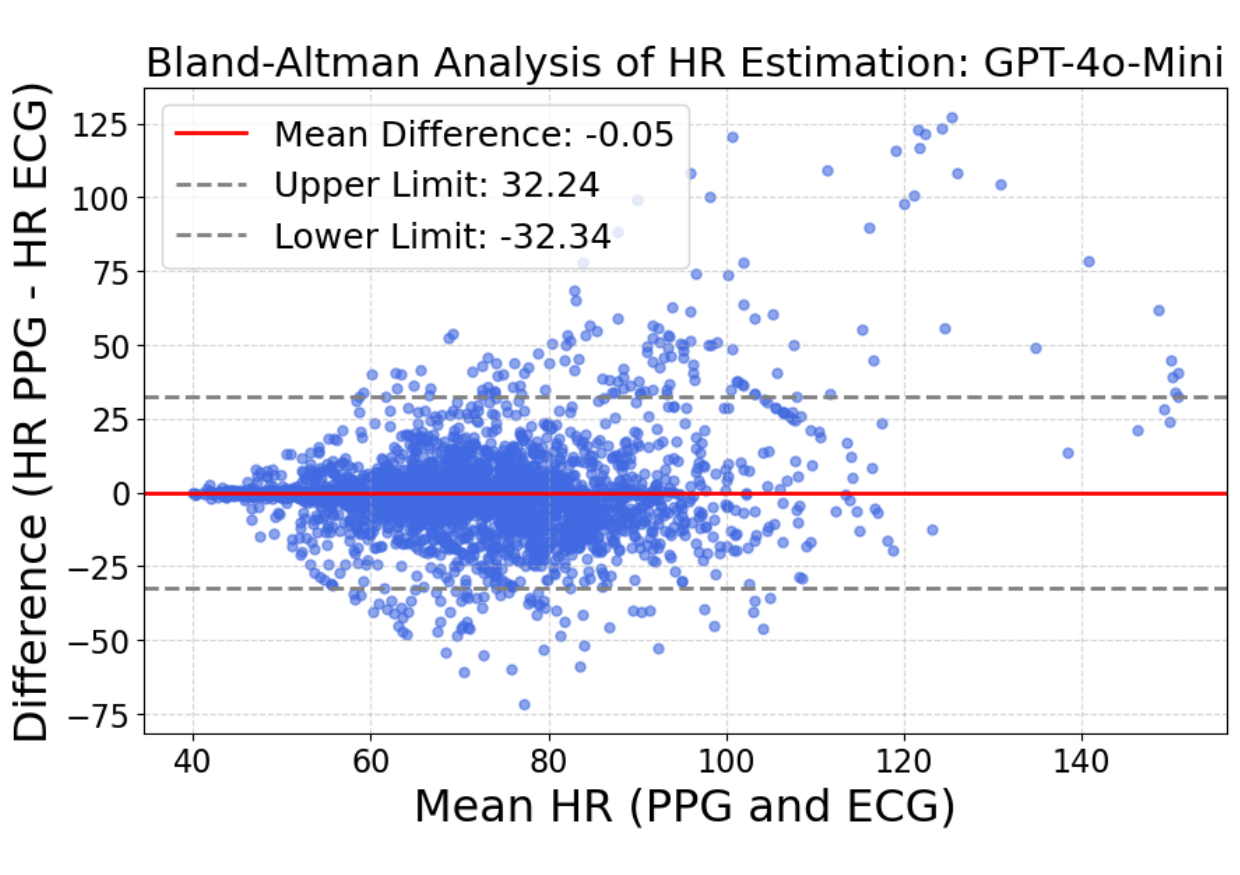}
    \caption{GPT4o-Mini}
    \label{fig:bland-4o-mini}
  \end{subfigure}
  \hfill 
  \begin{subfigure}[b]{0.36\textwidth}
    \includegraphics[width=\textwidth, trim=0cm 0.5cm 0cm 0.5cm, clip]{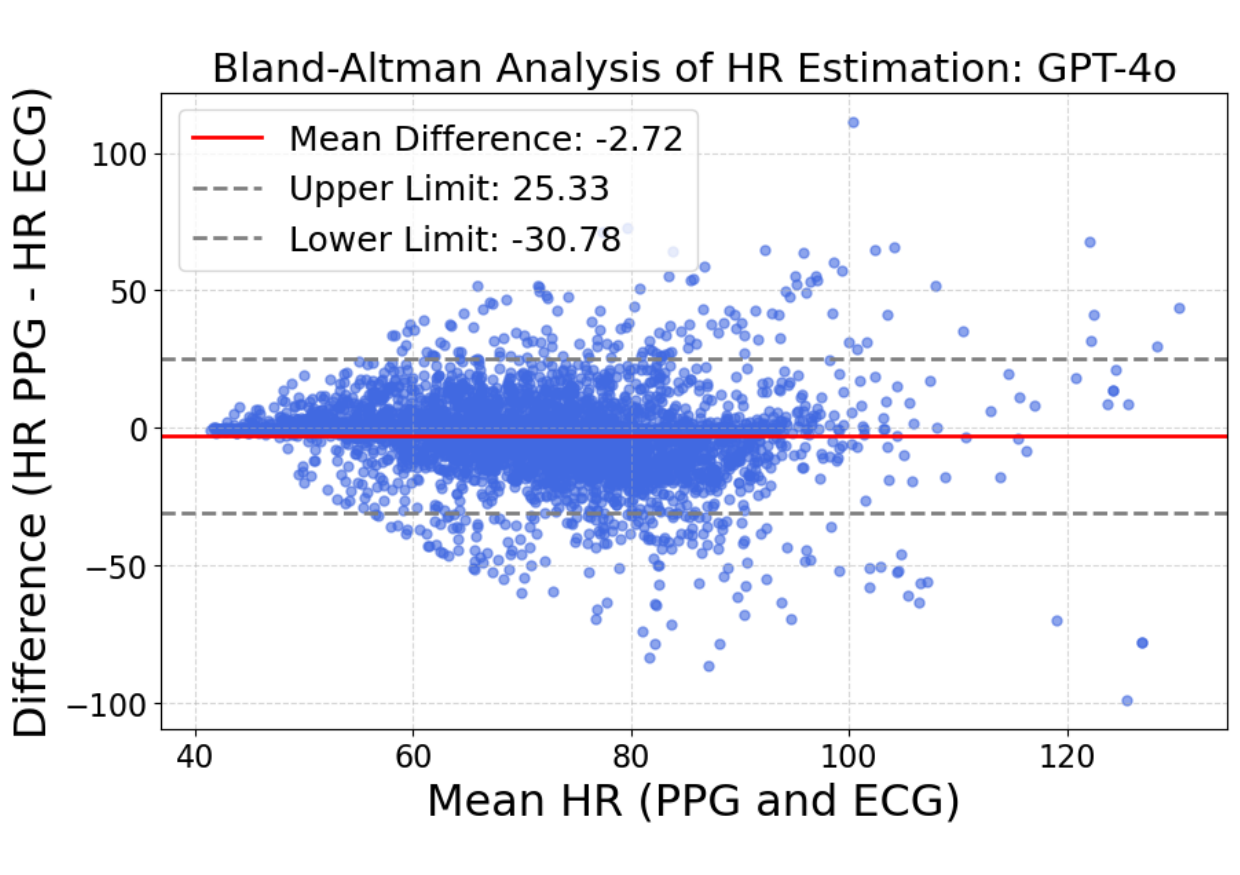}
    \caption{GPT4o}
    \label{fig:bland-4o}
  \end{subfigure}

  \caption{Bland-Altman analysis plots of the proposed agent and benchmark OpenAI models}
  \label{fig:bland-plots}
\end{figure}

In addition, Figure \ref{fig:bland-plots} demonstrates the Bland-Altman plots for our agent and OpenAI models. In these plots, the red line represents the mean difference, indicating bias, while the gray lines denote the 95\% limits of agreement, capturing the range within which most differences fall. Among the models, GPT-4o-Mini exhibits the lowest mean difference (-0.05), suggesting the least bias. However, in terms of the spread of errors, our agent demonstrates the narrowest range around the mean difference, with limits of agreement spanning from -12.03 to 6.86. In contrast, the error range is considerably wider for GPT-4o-Mini (-32.34 to 32.24) and GPT-4o (-30.78 to 25.33), indicating higher variability and lower agreement with the reference HR values.

Prior to evaluation, we removed estimated HR values outside the [40, 200] BPM range, as these were considered outliers. This threshold was chosen based on our dataset, which consists of healthy individuals aged 18 to 55, where HR values within this range are physiologically expected. Any HR estimates falling outside this range were excluded from the analysis. Table \ref{tab:outliers} summarizes the total number of extracted HR samples, the number of removed outliers, and the percentage of outliers for each method. As shown, GPT-4o produced the highest number of HR estimates (5,738), while GPT-4o-Mini and our agent extracted 4,075 and 2,479 samples, respectively. These differences in total extracted samples arise from the models' differing sensitivity to noise—since our prompts instructed the models to return NaN for noisy segments (see Figure {\ref{fig:responses}}). In terms of outlier removal performance, our agent demonstrated a significant advantage, identifying only 2 outliers (0.08\%). In contrast, GPT-4o-Mini exhibited an outlier rate of 17.55\% (715 outliers), and GPT-4o had a rate of 6.20\% (356 outliers). These results highlight our agent's superior robustness and reliability in HR estimation, with minimal outlier predictions compared to OpenAI models.

\begin{table}[!t]
\centering
\caption{Comparison of outlier removal across different methods. Outliers are estimated HR values outside the range [40, 200] BPM.}
\begin{tabular}{lccc}
\hline
\textbf{Method}   & \textbf{Total} & \textbf{Removed} & \textbf{Outlier \%} \\ \hline
GPT-4o-Mini       & 4075           & 715              & 17.55               \\
GPT-4o            & \textbf{5738}  & 356              & 6.20                \\
\textbf{Proposed} & 2479           & \textbf{2}       & \textbf{0.08}       \\ \hline
\end{tabular}
\label{tab:outliers}
\end{table}

The higher outlier rates in the OpenAI model approaches are due to the inherent uncertainty in LLM-generated and executed codes. Although Assistants execute codes in a controlled environment, the accuracy of the analysis depends on the correctness of the code generated. In many cases, the OpenAI models produced incomplete or inaccurate signal processing codes, which led to outliers in HR estimation. In contrast, our orchestrated agent employs validated and reliable processing tools, ensuring robust analysis and physiologically realistic results.
 



\section{CONCLUSIONS}
LLMs face limitations in token constraints, analytical rigor, and inefficiency in processing physiological time-series data. In this paper, we developed an LLM-powered agent to generate accurate health insights from physiological time-series by integrating well-established analytical tools. Built on the OpenCHA framework, our agent features an orchestrator that integrates user interaction, data sources, and analytical models. To demonstrate its effectiveness, we implemented a case study on HR estimation from PPG signals. The agent was evaluated on a PPG dataset from a remote health monitoring study and compared against OpenAI models. Experimental results show that our agent significantly outperforms benchmark models, achieving lower error rates and more reliable HR estimations. In future work, we plan to extend our agent to estimate additional physiological parameters, such as heart rate variability (HRV) and respiration rate. Additionally, we aim to benchmark our approach against more LLMs, including Gemini and PaLM. Furthermore, we will explore advanced prompt engineering and fine-tuning techniques to further improve evaluation.









\bibliographystyle{IEEEtran}
\bibliography{references}

\end{document}